\documentclass[letterpaper, 10 pt, conference]{ieeeconf}  

\IEEEoverridecommandlockouts                              

\usepackage{cite}
\usepackage{amsmath,amssymb,amsfonts}
\usepackage[ruled,linesnumbered]{algorithm2e}
\usepackage{algorithmic}
\usepackage{graphicx}
\usepackage{subfigure}
\usepackage{graphics}
\usepackage{textcomp}
\usepackage{xcolor}
\usepackage{mathptmx} 
\usepackage{times} 
\usepackage{bm}
\usepackage{multirow}
\usepackage{booktabs}
\usepackage{array}
\usepackage{tabularx,booktabs}
\usepackage{overpic}
\usepackage{url}
\usepackage[bookmarks=false]{hyperref}
\usepackage{rotating}
\usepackage{mathtools}
\usepackage{cite}
\usepackage{multirow}
\usepackage[T1]{fontenc}

\begin{document}

\title{\LARGE \bf Distributed Multi-Robot Obstacle Avoidance via Logarithmic Map-based Deep Reinforcement Learning}

\author{Jiafeng Ma $^{1}$, Guangda Chen $^{2}$, Yingfeng Chen $^{2}$, Yujing Hu $^{3}$, Changjie Fan $^{2}$ and Jianming Zhang $^{1}$
\thanks{
        This work is partially supported by Robotics Institute of Zhejiang University under Grant K12201.
}
\thanks{
        $^{1}$ College of control science and engineering, Zhejiang University, Hangzhou, Zhejiang, 310063, China. 
        {\small \{mjf\_zju, ncsl\}@zju.edu.cn.}
}
\thanks{
        $^{2}$ Fuxi Robotics in NetEase, Hangzhou, Zhejiang, China. {\small \{chenguangda, chenyingfeng1, fanchangjie\}@corp.netease.com.}}
\thanks{
        $^{3}$ Fuxi Lab in NetEase, Hangzhou, Zhejiang, China. {\small huyujing@corp.netease.com.}}
}

\maketitle

\begin{abstract}
        Developing a safe, stable, and efficient obstacle 
        avoidance policy in crowded and narrow scenarios for 
        multiple robots is challenging. Most existing studies either use 
        centralized control or need communication with other robots. 
        In this paper, we propose a novel logarithmic map-based 
        deep reinforcement learning method for obstacle avoidance 
        in complex and communication-free multi-robot scenarios. 
        In particular, our method converts laser information into a 
        logarithmic map. As a step toward improving training speed 
        and generalization performance, our policies will be trained in 
        two specially designed multi-robot scenarios. Compared to 
        other methods, the logarithmic map can represent obstacles 
        more accurately and improve the success rate of obstacle 
        avoidance. We finally evaluate our approach under a variety of 
        simulation and real-world scenarios. The results show that our 
        method provides a more stable and effective navigation solution for robots 
        in complex multi-robot scenarios and pedestrian scenarios. Videos are available at https://youtu.be/r0EsUXe6MZE.
\end{abstract}

\section{Introduction}
\label{sec:intro}

With the increasing application scale of mobile robots, the problem of multi-robot obstacle avoidance (MROA) becomes more and more important.
MROA problem requires each robot moves from one point to another while avoiding obstacles and other robots successfully. 
Different shapes of obstacles and other mobile robots make the problem even more challenging.

Methods for solving the MROA problem can be divided into two categories: centralized method and distributed method. 
For centralized methods \cite{wang2018personalized} \cite{alonso2017multi}, all mobile robots are controlled by a central server. 
These centralized control methods generate collision avoidance actions by planning optimal paths for all robots simultaneously through a global 
optimizer. 
But this kind of method is computationally complex and requires reliable synchronized communication between the central server and
all robots. When there is a communication problem or sensor failure of any single robot, the whole system will fail or be seriously disturbed. 
In addition, it is difficult to deploy centrally controlled multi-robot systems in unknown environments, e.g., in a workshop with human co-workers.

Different from the centralized method, the distributed method enables each robot to make appropriate decisions through its own controller. 
Furthermore, distributed methods can be divided into two groups: the traditional method and the learning-based method. 
Many traditional methods make decisions based on the principles of robot kinematics through velocity, acceleration, 
and other environmental information. Unlike traditional methods, learning-based methods map diverse sensor data like laser and 
images to robot's action through a learnable model such as a neural network.

Exiting traditional methods, such as \cite{fiorini1998motion,van2008reciprocal,berg2011reciprocal}, are reaction-based method. 
They specify a one-step interaction rule for the current geometry configuration.
The velocity obstacle (VO) \cite{fiorini1998motion} method plans a collision-free real-time trajectory based on the position of 
surrounding objects and the speed of other robots. Some of its excellent variants, such as reciprocal velocity obstacle (RVO) \cite{van2008reciprocal} 
and optimal reciprocal collision avoidance (ORCA) \cite{berg2011reciprocal}, are widely used in robot dynamic obstacle avoidance and game character 
obstacle avoidance.
Although these methods have achieved good results in solving the MROA problem, they also have some shortcomings. Firstly, they need real-time status of 
other robots, 
Secondly, the perfect and real-time sensor information is also required.

Deep reinforcement learning (DRL) has made extraordinary achievements in many fields, like Alpha zero in Go game \cite{silver2017mastering} and 
Open-AI Five in MOBA games \cite{berner2019dota}. 
A typical learning-based method for the MROA problem is the deep reinforcement learning (DRL) based method.
Although there have been some admirable learning-based works \cite{chen2017decentralized,chen2017socially}, 
those works still require the movement data of nearby robots and obstacles. 
Exiting DRL-based methods can be roughly divided into two classes, i.e., sensor-level method and map-based method. 
The sensor-level method \cite{fan2020distributed} is limited to specific sensor data. 
The map-based method \cite{chen2020distributed} uses a grid map that can be easily generated by 
using multiple sensors or sensor fusion.

However, due to the limited computing resources, the grid map-based method uses a lower resolution to represent the whole environment, which will 
lead to the loss of some important obstacle information when the obstacles are close. For example, it may mark the grid with obstacles as idle. Although 
the resolution of the grid map can be very high when there are enough computing resources, it will also increase the state space, resulting in slower 
convergence.

In this paper, we propose a novel logarithmic map-based DRL method to handle the MROA problem. We use the down-sample skill mentioned in 
\cite{liu2020robot,zhang2021ipaprec}, 
to reduce the dimension of the 2D laser information. Then we propose a logarithmic transformation method to process the laser information. 
At the same resolution as the grid map \cite{chen2020distributed}, our method uses more pixels to represent the near information. We use the logarithmic graph to help the robot pay more attention to the surrounding environment and learn stable and 
efficient obstacle avoidance strategies faster and better.
Besides, our approach doesn't need any communication with other mobile robots, it only needs its own sensor data and the relative posture to the goal 
point. 
We apply distributed Proximal Policy Optimization (DPPO) to train a neural network that maps high-dimensional observations to low-dimensional 
actions.
We train and evaluate our method in different complex scenarios. Our contributions can be summarized as the following points:
\begin{itemize}
	\item We propose a logarithmic map-based multi-robot obstacle avoidance approach in communication-free scenarios, 
        where the logarithmic map can help the robot pay attention to the nearest information and avoid obstacles in complex scenarios efficiently.
	\item We deployed our model in a real robot and demonstrated the practical effect of our proposed 
        obstacle avoidance method although there are many interference factors in the 
        actual environment.
\end{itemize}

The rest of this paper is organized as follows. Related works will be introduced in section \ref{rw}. 
In section \ref{approach}, our work will be discussed in detail. Section \ref{exp} presents the simulation experiment results and the real-world experiments,
followed by conclusions in section \ref{conclusion}.
\section{RELATED WORKS}
\label{rw}

\subsection{Traditional multi-robot obstacle avoidance}
Fiorini et al. \cite{fiorini1998motion} propose an algorithm that directly predicts potential collisions in time-varying environments from velocity information which is the first-order algorithm.
This method exploits the concept of Velocity Obstacle (VO) to map the dynamic environment into the robot's velocity space. It excludes the speed that may cause a robot collision in the future, but when two moving objects are about to collide,  the object will shake back and forth. To solve this unstable phenomenon, RVO \cite{van2008reciprocal} assumes that another agent is also using the same decision as same as a current agent.  Besides, ORCA \cite{berg2011reciprocal}, an optimized version of ROV, is one of the most successful methods for solving the multi-robot navigation problem.
By introducing a time window, the relative position is converted into velocity, so that the optimal velocity can be calculated in the same velocity coordinate system. ORCA needs to ensure that the calculated speed is the optimal solution for both parties to calculate the global optimal solution. This method can adapt to large-scale robot system easily. \cite{snape2010smooth,alonso2018cooperative} are other excellent improved versions of the previous method. 

Although these approaches have achieved some success, they require some assumptions about the actual environment, precise environmental information, and other robots' real-time information.
Our method does not need to communicate with other robots and can learn the optimal policy directly based on the sensor information and the relative pose of the target point.
\subsection{Learning based multi-robot obstacle avoidance}
With the rapid development of deep learning, learning-based methods are becoming more and more popular. 
Deep learning provides new solutions for many problems. Qin et al. \cite{qin2021deep} propose an effective imitation learning-based path 
planning system that takes the sensor data as well as pedestrians' dynamic information as inputs, enabling socially compliant 
navigation in dynamic pedestrian environments. But it requires a large amount of pre-prepared data, which is very time-consuming and 
labor-intensive to collect.

In addition to imitation learning, DRL performs well in solving collision avoidance problems. Tai et al. \cite{tai2017virtual} use DRL to learn an optimal policy that navigates the robot from one point to another without a map.
It only uses a 10-dimensional laser and the relative position of the target. However, this method is not suitable for dynamic and 
complex environments as it obtains too little environmental information. 
Chen er al. \cite{chen2017decentralized} propose a DRL-based method to solve the MROA problem. It is a kind of distributed sensor-level approach but still needs other agents' information. A sensor-level method proposed in \cite{fan2020distributed} uses three consecutive frames of radar data directly as the main 
input and applies their model in real robots. However, their method can only use high-dimensional Lidar data as input, and its performance in complex environment is ordinary. Chen et al. \cite{chen2020distributed, chen2021deep} propose a map-based approach, which represents environmental features with a local grid map, that can be generated by different sensors. It has better performance than the sensor-level method but its effect depends largely on the resolution of the grid map, but the higher the resolution, the greater the demand for computing resources.
When the resolution is low, the grid map can't completely represent the environment, which will make a great error between the environment represented by the grid and the real environment. 

In this paper, we use the logarithmic map to represent the environment information around the robot, we extract laser information at different 
distances and angles in the form of concentric rings and represent it as a two-dimensional matrix. The state represented by the logarithmic 
map is more realistic. The closer it is to the center, the more complete the information is.
\section{Approach}
\label{approach}

In this section, we first describe our method of logarithmic map generating. After that, we briefly describe our mobile robot obstacle 
avoidance problem from the perspective of reinforcement learning, and then introduce the network structure and training details.  

\begin{figure}
	\centering
        \subfigure[raw laser data]{
                \label{fig1a}
                \includegraphics[ width=0.3\linewidth]{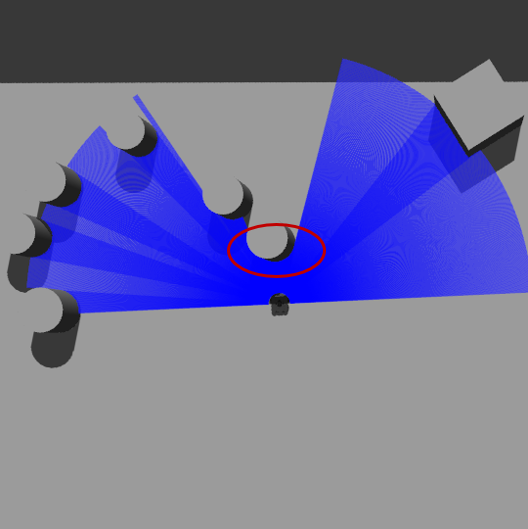}}  
        \subfigure[grid map]{
                \label{fig1b}
                \includegraphics[ width=0.3\linewidth]{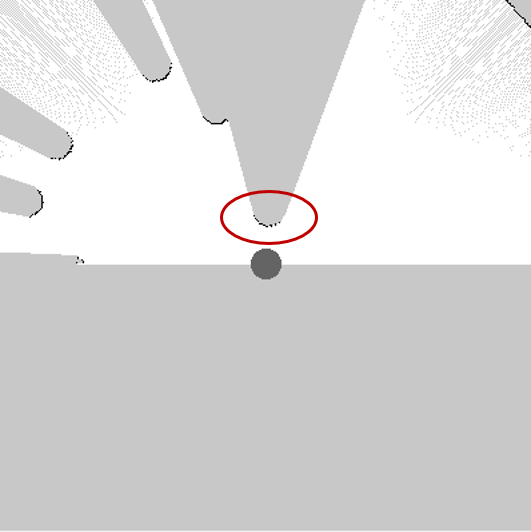}}
        \subfigure[logarithmic map]{
                \label{fig1c}
                \includegraphics[ width=0.3\linewidth]{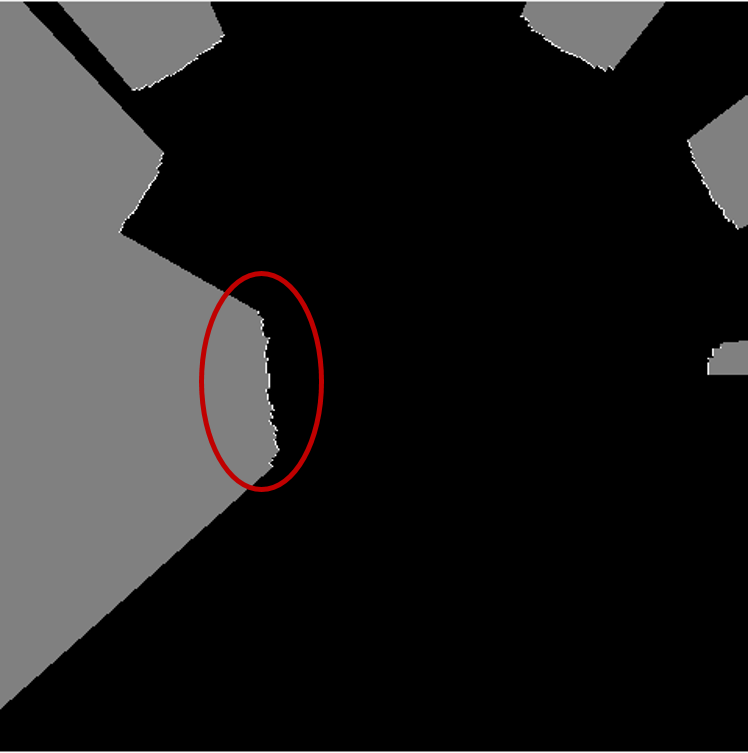}}   
	\caption{In all figures, the same obstacle is in the red circle, but the logarithmic map can have a higher resolution. 
	Besides, in the logarithmic map, obstacles of the same size but different distances also have different resolutions, where the obstacles near are more obvious than those in the distance.}	\label{fig1}
\end{figure}

\subsection{Logarithmic map generation}
Fig.~\ref{fig1} shows the difference between raw laser date, grid map\cite{chen2020distributed} and logarithmic map. The process of logarithmic graph generation can be specified as follows:
First, we use the down-sampling skill to reduce the 
dimension of the raw radar information, $o_{ds}^t=[\min(o_{s_{0:a}}^t),\min(o_{s_{a:2a}}^t),\cdots]$
where $min(o_{s_{a:2a}}^t)$ means we take the minimum of every $a$ degrees' laser 
scans as the generalized representation of the corresponding radar information.
Secondly, while some papers \cite{chen2020distributed,yao2021crowd,liu2020robot} use the traditional skill (Fig.~\ref{fig1b}) to process the raw laser scans, 
we find that the grid map can not distinguish the importance of different distances, but we know that the closer the more essential.
Our method uses concentric rings of a certain width to intercept a circle of radar and straighten it into an one-dimensional vector according 
to its grid map-like representation: $0.0$ for free place, $1.0$ for obstacle and $0.5$ for unknown area. Then we introduce the implementation of our 
transformation. We set $n_s$ the sample number, $O_{s,max}$ the max range of laser, 
$o'_s$ is the distance of each laser in $O_s$, our transformation equation is \begin{equation}f(x)=e^x-1, \end{equation} we can calculate interception intervals $g$ by
\begin{equation}g=\frac{log(O_{s,max}+1)}{n_s},\end{equation}
the width of ring can be defined as 
\begin{equation}w=(f(g\times k),f(g\times (k+1))), k=[0,1,2,\dots,n_s-1],\end{equation}
where $f(g\times k)$ is the lower bound and can be included; $f(g\times (k+1))$ is the upper bound but cannot be included. We set $1.0$ while $w$ includes $o'_s$, 
$0.0$ while the upper bound smaller than $o'_s$ and $0.5$ while the lower bound bigger than $o'_s$. Finally, the data will be transformed from polar coordinates to Cartesian coordinates by the linear-polar inverse transformation tool in OpenCV. After that, we will get the final logarithmic map ($o_{l,i}^t$) in Fig.~\ref{fig1}(c). In this paper, we set $n_s=48$ and $O_{s,max}=6.0$ meter.

\subsection{Reinforcement learning set up}
Multi-robot obstacle avoidance requires $N$ independent robots in the same environment. 
Each robot $i$ ($0 \leq i < N$) can successfully move from the current position to the target point without collision through 
its own obstacle avoidance strategy. At time step $t$, robot $i$ receives it's laser information ($o_{s,i}^t$), 
relative angle ($\theta_i^t$) and distance ($d_i^t$) to the target point.
Then it takes an action ($a_{\pi_i}^t$) by its policy ($\pi_i$).

The problem can be seen as a partially observed Markov decision process (POMDP) which is defined by a tuple ($S, A, P, R,\Omega, O$), where $S$ is the state space, $A$ is the action space, $P$ is the transition probability from the previous state to the next state, 
$R$ is the reward function that can be manually engineered or learned through some special methods, 
$\Omega$ is the observation space that is different from classic MDP, 
and $O$ is the observation function that captures the relationship between the state and the observations (and can be action-dependent). 

In this paper, we use distributed Proximal Policy Optimization (DPPO) algorithm to train our agent. 
DPPO is an extended version of the Proximal Policy Optimization (PPO) algorithm \cite{schulman2017proximal} which uses multiple robots to collect trajectories at the same time in different training environments. 
The goal of reinforcement learning \cite{wiering2012reinforcement} is to learn an optimal policy of the agent $\pi_* (a,s)=p_* (a|s)$ that maximizes the expectation of the cumulative discounted rewards. We use generalized advantage estimator (GAE) \cite{schulman2015high} while calculating the objective of the agent's policy. 
The objective function $J$ is
\begin{equation} J(\theta) = \sum_{t=0}^{T}\textrm{min}(\frac{\pi _\theta (a,s)}{\pi _{\theta_{old}} (a,s)}A,  \textrm{clip}(\frac{\pi _\theta (a,s)}{\pi _{\theta_{old}} 
(a,s)}, 1-\varepsilon , 1+\varepsilon )A), \end{equation}
$A$ is the advantage calculated by GAE, 
$\varepsilon$ is the clip function ration and $\theta$ is the parameter of the policy. The key components of our algorithm are as follows:
\subsubsection{Observation space}
We set every robot $i$ can only get two types of observation information at each time 
step $t$, $O_i^t=[o_{s,i}^t,o_{p,i}^t]$, where $o_{s,i}^t$ is the radar information for 180 degrees, 
$o_{p,i}^t=[x_i^t,y_i^t,\phi_i^t]$ is the relative posture of the robot and the target point. 
After we map the radar information to a logarithmic map $(o_{l,i}^t)$, the final observation is  $[o_{l,i}^t,o_{p,i}^t]$. 
\subsubsection{Reward design}
The reward function of all robots in the training scenarios are independent and the same. 
The objective of the reward function is to enable our model to learn a policy which can help robots successfully avoid obstacles 
and reach local targets points in multi-robot scenarios. The rewards are designed as follows:
\begin{equation}r^t=r_a^t+r_c^t+r_d^t+r_s^t,\end{equation}
\begin{equation}r_a^t=\left\{\begin{matrix} 
        r_{arrive} \quad if \ arrive, \\  
        0 \qquad otherwise,
      \end{matrix}\right. \end{equation}
\begin{equation}r_c^t=\left\{\begin{matrix} 
        r_{collision} \quad if \ collision,\\ 
        0 \qquad otherwise,
      \end{matrix}\right. \end{equation}
\begin{equation}r_d^t=\tau (||p^{t-1}-p_g||)-||p^t-p_g||,\end{equation}
\begin{equation}r_s^t=r_{step},\end{equation}
where $p^t$ is the current position of robot, $p_g$ is goal's position. 
We use $d_c$ to denote the distance between robot and obstacles, 
$d_{gmin}$ and $d_{cmin}$ to denote the minimum distance to goal and obstacle. 
When $||p^t-p_g ||<d_{gmin}$, the robot arrives.  
$r_{collision}$ specifies the penalty when the robot encounters a collision. 
$r_d^t$ encourages the robot to move towards the target point and punishes the behavior away from the target point. 
$r_s^t$ help the robot reach the target point with fewer steps.
We set $r_{arrive}=500$, $r_{collision}=-500$, $r_{step}=-5$, $\tau=200$.

\subsubsection{Action space}
The action of each robot $i$ at time step $t$, 
including linear velocity ($v_i^t$) and angular velocity ($\omega_i^t$), 
is limited to a fixed range. 
In this paper, we use continous action space and set $v_i^t\in[0,0.6]$ (in meters per second), 
$\omega_i^t\in[-0.9,0.9]$ (in radians per second).

\begin{figure}
	\centering
	\includegraphics[ width=0.9\linewidth]{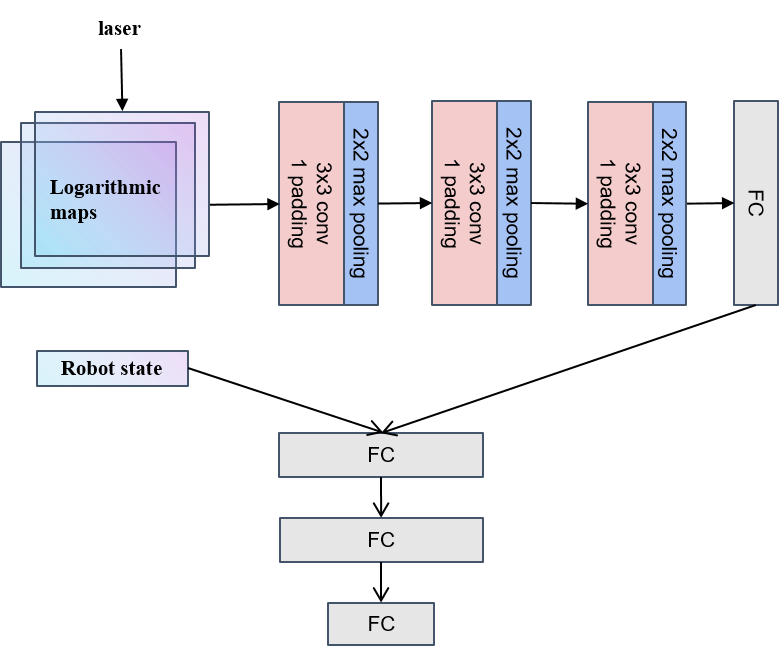}
	\caption{The network structure of our reinforcement learning algorithm.}	\label{fig2}
\end{figure}

\subsection{Network Architecture}

Fig.~\ref{fig2} shows the framework of our policy network, our value network also uses the same network structure. 
We use three frames logarithmic map which is mentioned in section \ref{approach} and relative goal position as input. 
The last fully-connected layer outputs two-dimensional vectors as our action space is continuous.

Specially, we first process high-dimensional laser inputs through three convolutional layers, 
each layer is followed by a maximum pool layer. All convolutional layers convolve filters with kernel size = 3, 
stride = 1 over the three input logarithmic maps and applies ReLU nonlinearities \cite{nair2010rectified}. 
The output of the last convolutional layer is connected to a fully connected layer with 512 units. 
Then we concatenate the output of the fourth layer with relative posture together as the input of the next fully connected layer followed 
by another fully connected layer with 512 units. The last layer output the mean of linear velocity and the mean of angular velocity. 
For continuous action space, the actions are sampled from the Gaussian distribution whose log standard deviation is generated by 
a standalone network.

\subsection{Training}

\begin{figure}
	\centering
        \subfigure[Crowd scenario]{
                \label{fig3a}
                \includegraphics[ width=0.3\linewidth]{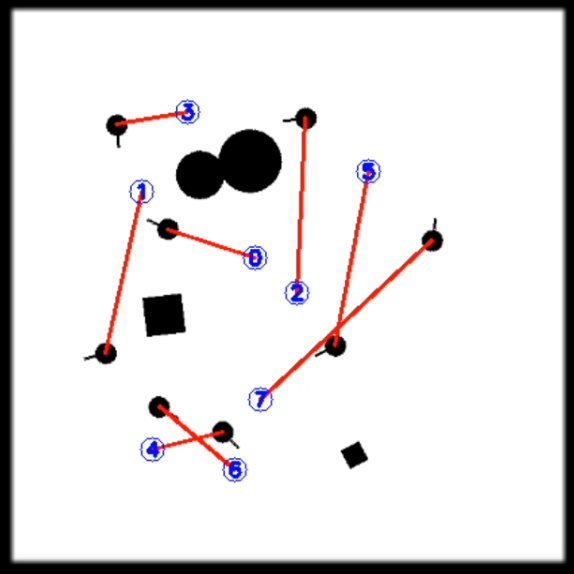}}   
        \subfigure[Circle scenario]{
                \label{fig3b}
                \includegraphics[ width=0.3\linewidth]{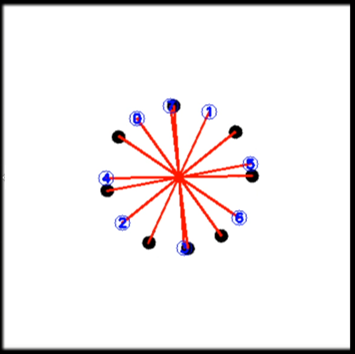}}   
        \subfigure[Narrow scenario]{
                \label{fig3c}
                \includegraphics[ width=0.3\linewidth]{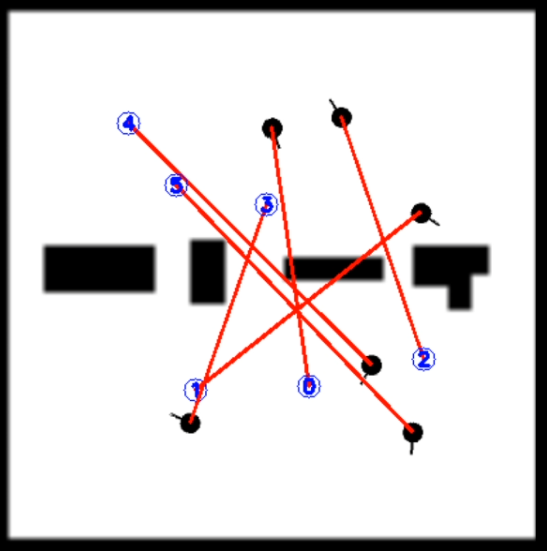}}     
        \quad
	\caption{Three scenarios for training. The blue numbers represent the goal position of one robot, red line is the straight path to goal. 
        Black objects other than robots are obstacles.}	\label{fig3}
\end{figure}

The simulation environment we use is specially designed for laser-based navigation$^{1}$. 
It supports the simultaneous training of multiple robots in different scenarios. 
Fig.~\ref{fig3} shows three types of scenarios used in our training phases. The environment in Fig.~\ref{fig3a} is designed for robots to learn good obstacle avoidance strategies. The other two environments are designed for robots to learn strategies in special scenarios.

\footnotemark[1]{
$https://github.com/DRL-Navigation/img\_ env.$
}

To prove the effectiveness of our method in various complex scenarios, our training strategy has the following
special points: 
\begin{itemize}
        \item We train our strategies in multiple scenarios in parallel, which brings robust performance.
        \item Two combination scenarios (Comb1 and Comb2) are designed to train our strategies. Comb1 consists of a crowd scenario and a circle scenario, Comb2 consists of a crowd scenario and a narrow scenario. Comparing the policies trained in different environments can better prove the effectiveness of our method.
        \item The shape and size of obstacles in all scenarios are random. 
        \item The starting point and target point of each robot are random in a certain range.
        \item A simple two-stage learning is applied while robots are trained in Comb2, we set further target points as the number of epochs increases.
\end{itemize}

\begin{figure}
	\centering
        \subfigure[Comb1]{
                \label{fig4a}
                \includegraphics[ width=0.9\linewidth]{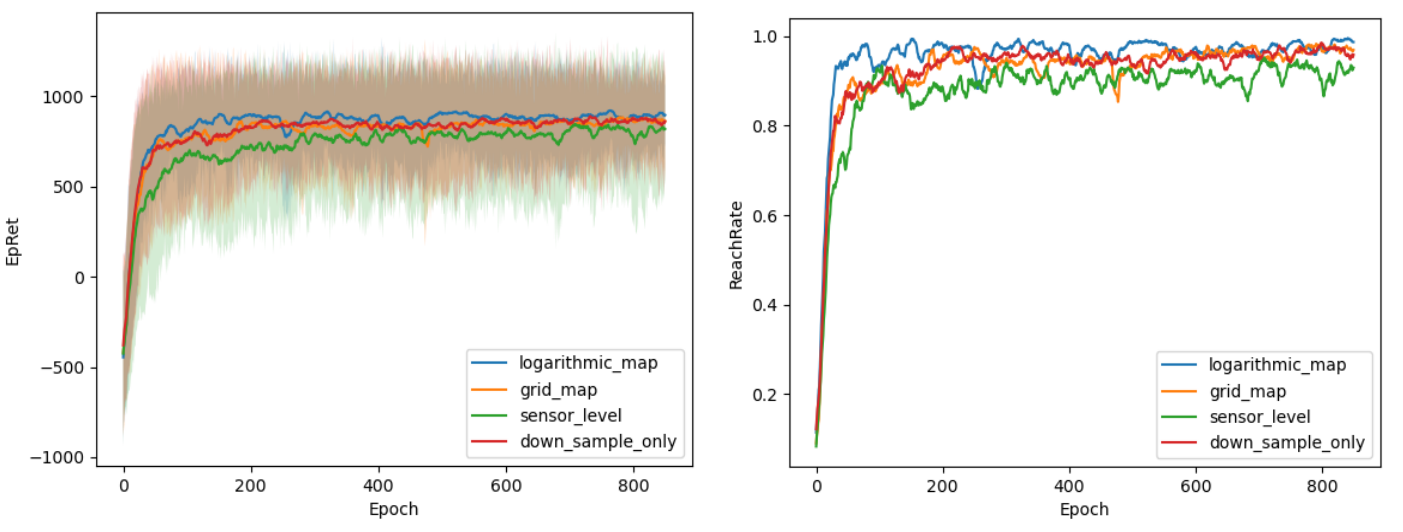}}  
        \\
        \subfigure[Comb2]{
                \label{fig4b}
                \includegraphics[ width=0.9\linewidth]{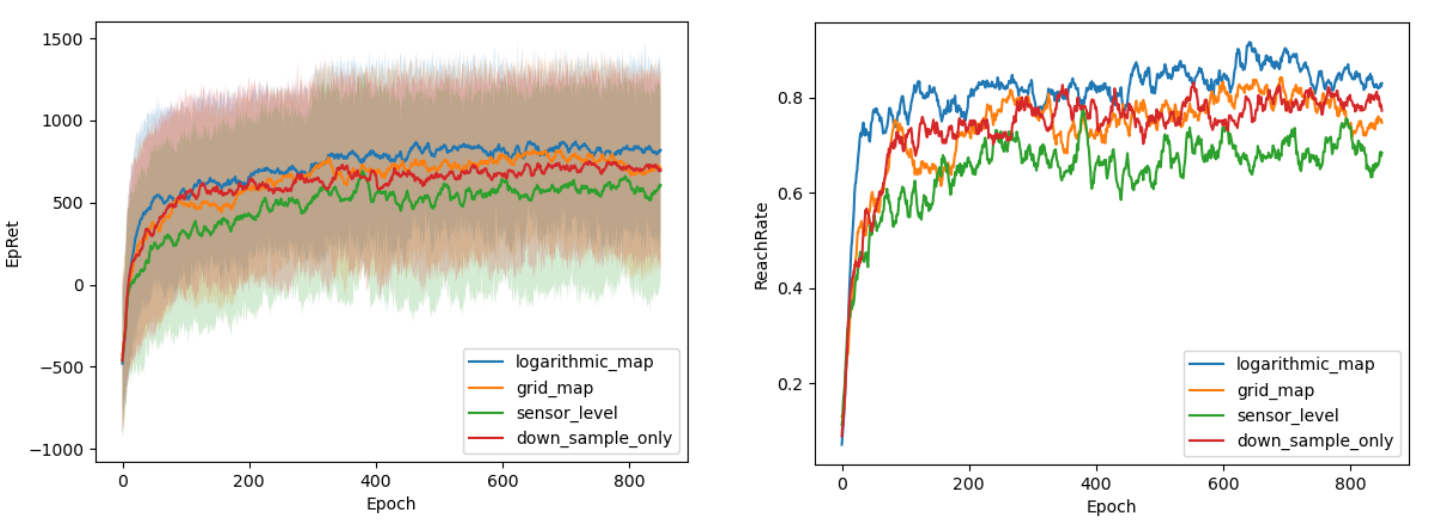}}      
        \quad
	\caption{The training result of two combination scenarios, where EpRet represents the average reward of each epoch and ReachRate is the arrive rate of each epoch.}	\label{fig4}
\end{figure}

Fig.~\ref{fig4} shows the training results, including average reward curves and reach rate curves of both Comb1 and Comb2. We test our model every 20 epochs for 20 epochs while training to get the reach rate.
It can be found that, in both two scenarios, our method converges faster and has the highest reward and success rate. Especially in Comb2, 
\section{Experiments}
\label{exp}

In this section, we evaluate our method in both simulators and the real world. 
Firstly, we introduce the main details of our method, including the hyper-parameters, 
hardware, and software for training. Then we define the performance metrics in virtual environments 
and evaluate our method by comparing it with two comparative groups and one ablation group. 
Finally, we also deploy and verify our method on a real robot. The result shows that the policy we trained 
allows robots to perform better in different scenarios and it also works well in reality. 
The following are four experimental groups. The first two are the comparative experimental group, the third group is the current group, and the last group named angular-map-based is the ablation experimental group to illustrate the impact of down-sampling techniques on our method.

\begin{itemize}
        \item Sensor-level: The method proposed by \cite{fan2020distributed}. It uses the raw laser from Lidar as the main input. They process the information through several 1D convolutional layers.
        \item Map-based: The method proposed by \cite{chen2020distributed}. It uses the egocentric local grid maps as inputs. They process the information through several 2D convolutional layers.
        \item Logarithmic map-based: The method proposed in this paper. We use the logarithmic maps as inputs and process the information through several 2D convolutional layers.
        \item Angular-map-based: Only use down-sample skill mentioned in our method to process laser information, then process the information through several 1D convolutional layers.
\end{itemize}

Considering the fairness, we use the same training tricks as well as our method. In particular, the number of frames is three, 
the number of raw lasers is 960, the size of the local grid map and the logarithmic map is $(48, 48).$ 

\begin{table}
        \centering
        \caption{HYPER-PARAMETERS}
        \label{table1}
        \begin{tabular}{|c|c|}
                \hline
                Hyper-parameters & Value \\ \hline
                Learning rate for policy network & 0.0003 \\ \hline
                Learning rate for value network	& 0.001 \\ \hline
                Discount factory & 0.99 \\ \hline
                Steps per epoch & 2000 \\ \hline
                Robot radius & 0.17 \\ \hline
                Clip ratio & 0.2 \\ \hline
                Lambda value for GAE & 0.95 \\ \hline
        \end{tabular} 
\end{table}
\subsection{Training setup}

The hyper-parameters of our algorithm are listed in Table~\ref{table1}.  
The training of our policy is implemented in PyTorch. The training hardware is a desktop PC with one i7-10700 CPU and one NVIDIA RTX 2060 super GPU. All methods are trained for about 800 epochs to ensure convergence. The model with the highest reach rate will be saved during training.
\begin{figure}
	\centering
        \subfigure[Crowd and narrow scenarios]{
                \label{fig5a}
                \includegraphics[ width=0.83\linewidth]{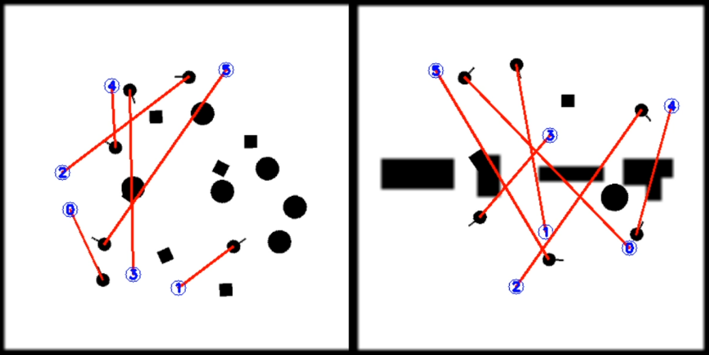}}  
        \\
        \subfigure[Circle1 scenario]{
                \label{fig5b}
                \includegraphics[ width=0.4\linewidth]{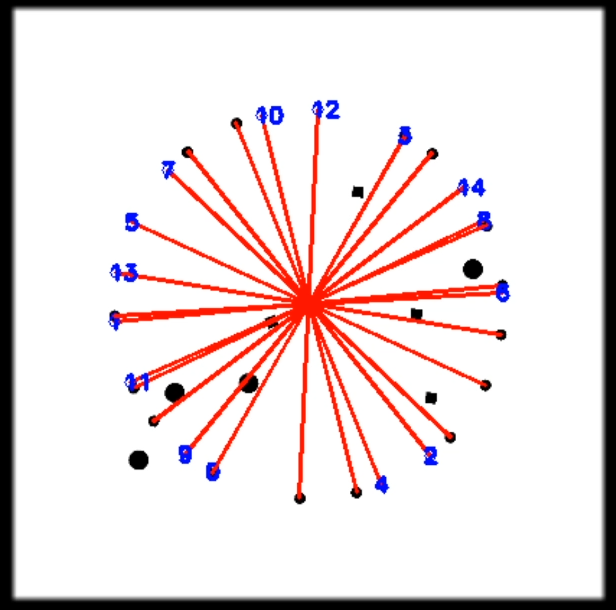}}      
        \subfigure[Circle2 scenario]{
                \label{fig5c}
                \includegraphics[ width=0.4\linewidth]{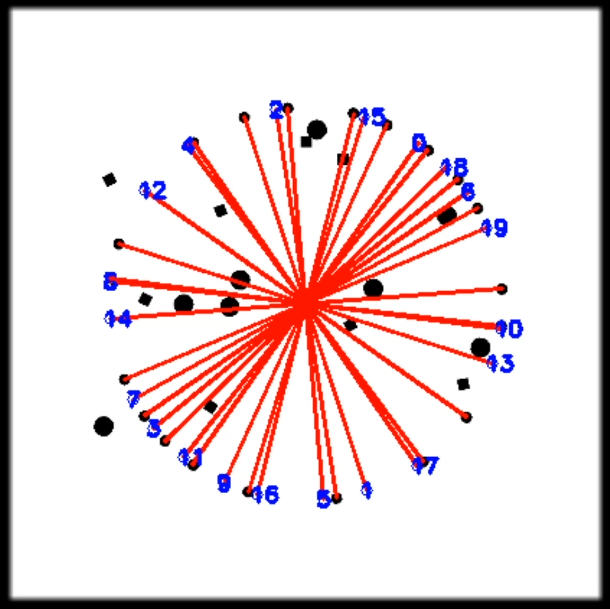}}    
        \subfigure[Cross scenario]{
                \label{fig5d}
                \includegraphics[ width=0.4\linewidth]{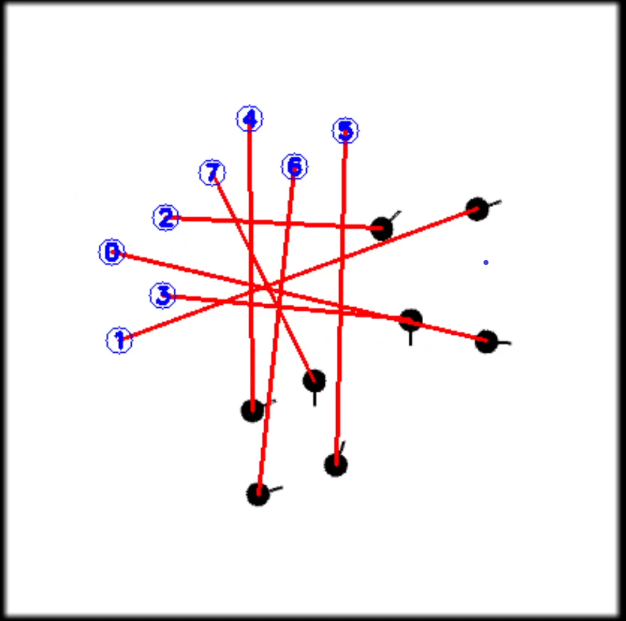}}    
        \subfigure[Corridor scenario]{
                \label{fig5e}
                \includegraphics[ width=0.4\linewidth]{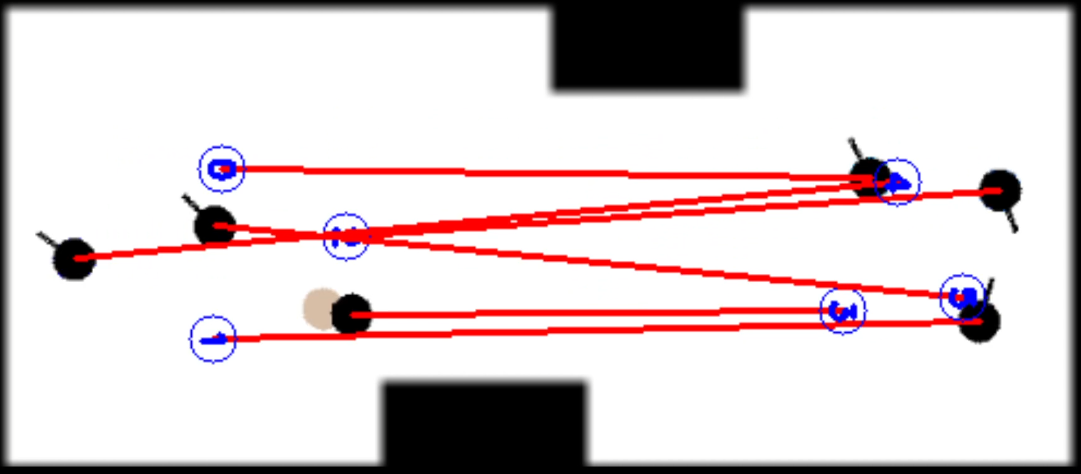}}    
        \quad
	\caption{Scenarios for testing}	\label{fig5}
\end{figure}
\subsection{Simulation Experiments}
The performance metrics we use are as follows:
\begin{itemize}
        \item Arrive rate (Ar): the ratio of the episodes ending with
        robots reaching their targets without collision. It indicates the stability of our robots.  
        \item Average angular velocity (Aav): the average angular velocity of all robots during the test.
        \item Average trajectory distance (Atd): the average distance all robots cost from start point to endpoint without collision. It indicates the efficiency of robots.
\end{itemize}

To verify the effect of our method, we test our method and other approaches in a variety of scenarios in Fig.~\ref{fig5}. 
Table~\ref{table2} summarizes the detailed test result. We test 50 epochs in each scenario.

\subsubsection{Crowd and narrow scenarios}
\begin{figure*}
	\centering
        \subfigure[Trajectories of logarithmic map-based method]{
                \label{fig6a}
                \includegraphics[ width=0.2\linewidth]{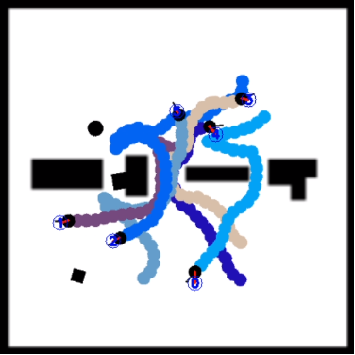}}  
        \subfigure[Trajectories of sensor-level method]{
                \label{fig6b}
                \includegraphics[ width=0.2\linewidth]{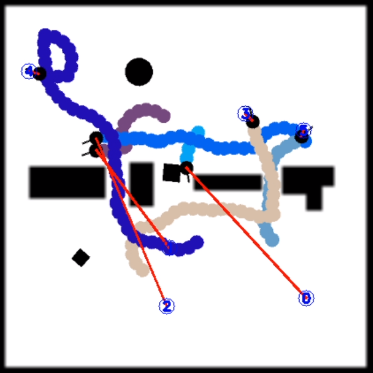}}  
        \subfigure[Trajectories of map-based method]{
                \label{fig6c}
                \includegraphics[ width=0.2\linewidth]{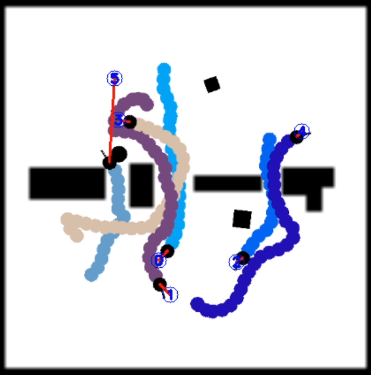}}  
        \subfigure[Trajectories of angular-map method]{
                \label{fig6d}
                \includegraphics[ width=0.2\linewidth]{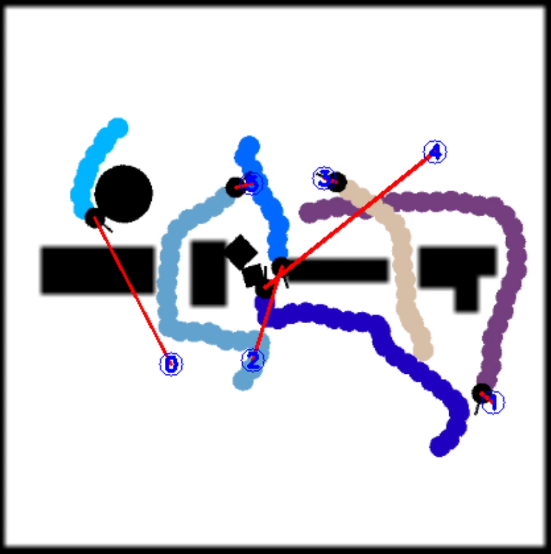}}  
	\caption{Trajectories while testing policies in narrow scenario}	\label{fig6}
\end{figure*}

Fig.~\ref{fig5a} used to test the model we trained in Comb2, two scenarios both have six robots with random positions. 
The robots in the left scenario need to avoid 16 random obstacles to reach the random target point, and the robots in the right scenario need 
to pass through the narrow road and avoid 3 random obstacles to reach the random target point opposite the channel. The quantitative difference between our policy and other policy in Table~\ref{table2} is also verified qualitatively by the difference in their trajectories illustrated in Fig.~\ref{fig6}. In particular, our policy can quickly perceive the obstacles in the channel and choose another feasible way.

\subsubsection{Circle scenarios}
\begin{figure}
	\centering
        \subfigure[Trajectories in Circle1 scenario]{
                \label{fig7a}
                \includegraphics[ width=0.4\linewidth]{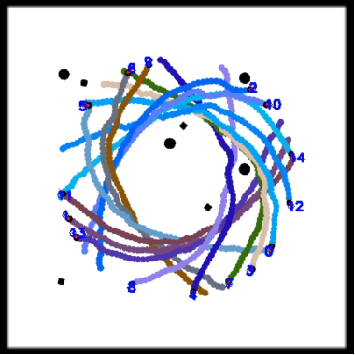}}  
        \subfigure[Trajectories in Circle2 scenario]{
                \label{fig7b}
                \includegraphics[ width=0.4\linewidth]{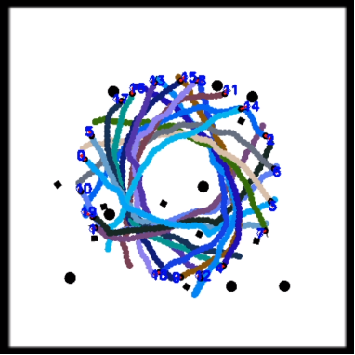}}  
	\caption{Trajectories while testing policies in circle scenarios}	\label{fig7}
\end{figure}

In circle scenarios, all robots form a circle and need to pass through random obstacles to reach the target point (the line between the target point and the robot crosses the center of the circle). We test the model we have trained in Comb1.
Fig.~\ref{fig5b} has 15 robots and 8 obstacles, Fig.~\ref{fig5c} has 20 robots and 16 obstacles with smaller circle radius range ([5,7] and [4,6] 
respectively). We use the scenario shown in Fig.~\ref{fig5c} because we find that in Fig.~\ref{fig5b}, the performance gap of each method is not obvious. The trajectories generated by our method are illustrated in Fig.~\ref{fig7}.

\subsubsection{Classical scenarios}
\begin{figure}
	\centering
        \subfigure[Trajectories of logarithmic map-based in crossing scenario]{
                \label{fig8a}
                \includegraphics[ width=0.4\linewidth]{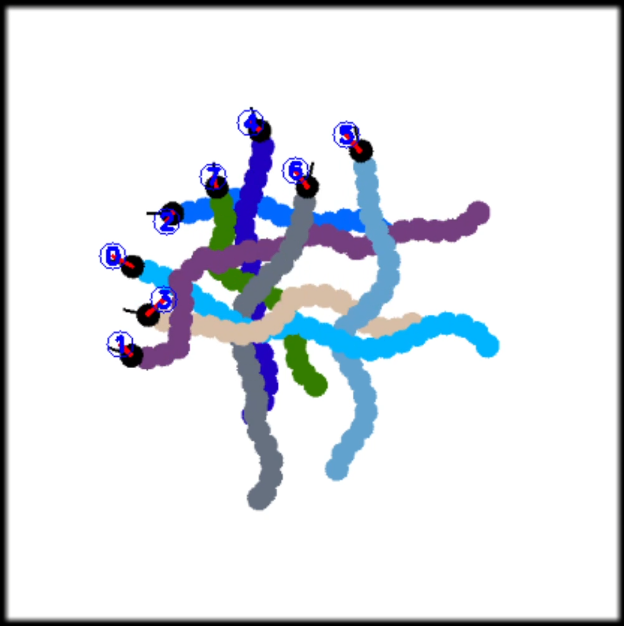}}  
        \subfigure[Trajectories of grid map-based in crossing scenario]{
                \label{fig8b}
                \includegraphics[ width=0.4\linewidth]{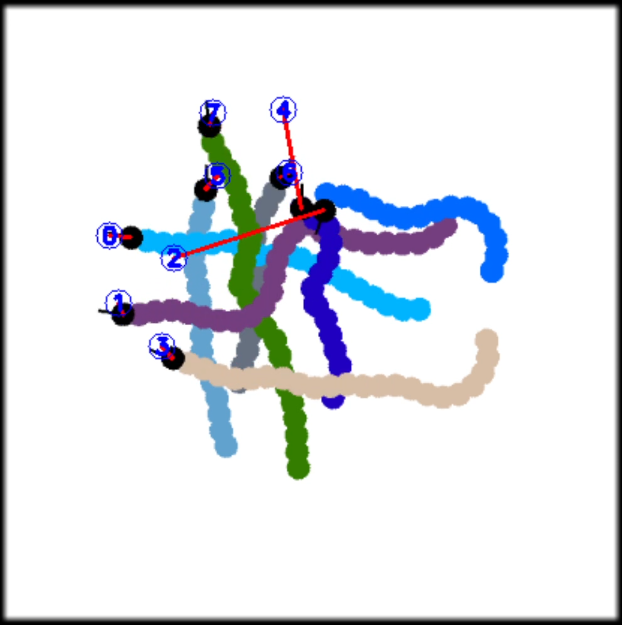}}  
        \subfigure[Trajectories of logarithmic map-based in narrow scenario]{
                \label{fig8c}
                \includegraphics[ width=0.4\linewidth]{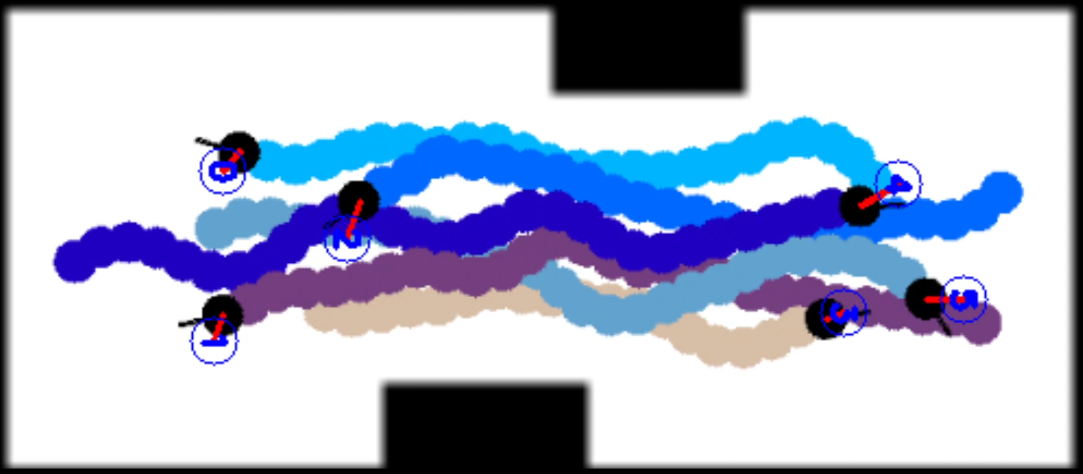}}  
        \subfigure[Trajectories of grid map-based in narrow scenario]{
                \label{fig8d}
                \includegraphics[ width=0.4\linewidth]{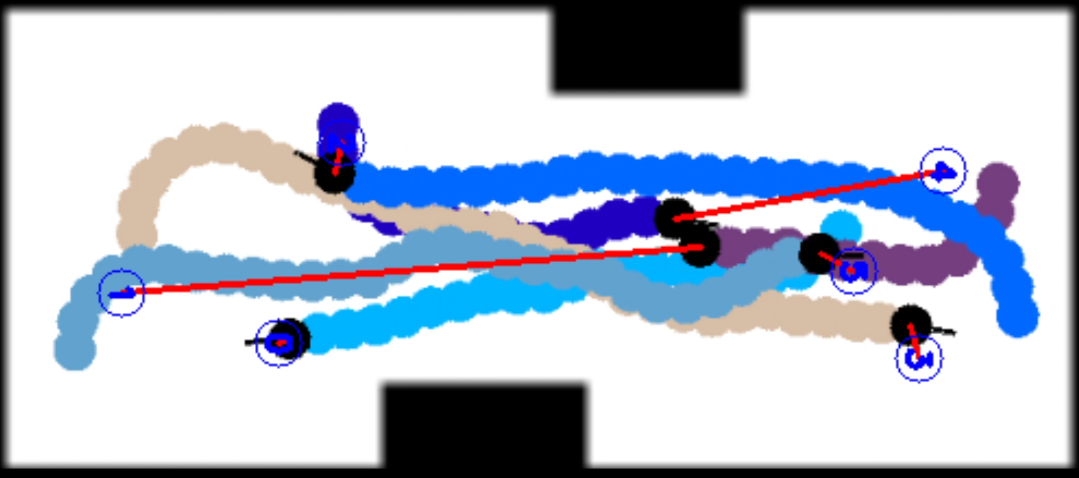}}  
        \quad
	\caption{Trajectories while testing policies in classical scenarios}	\label{fig8}
\end{figure}

Two classical scenarios will be applied to test our models trained in Comb1, namely the crossing scenario and the corridor scenario. The initial position and angle of each robot and the position of each target point are random in a certain range. As for the crossing scenario (as is shown in Fig.~\ref{fig5d}), robots are separated into two groups with 4 robots each, and their paths intersect in the center of the scenario. In the corridor scenario which is shown in Fig.~\ref{fig5e}, two groups (three robots in each group) exchange their positions via a narrow corridor connecting two open regions. We illustrate the trajectories of the grid map-based method and the logarithmic map-based method in Fig.~\ref{fig8}. Our method performs better although those scenarios have never been encountered during training.

According to the result, our method (logarithmic map-based) performs more stable, reliable, and efficient with the highest arrive rate and the lowest average angular velocity in all three scenarios, 
The ablation experiment also proves that the logarithmic map can greatly improve the obstacle avoidance ability of the robot. Please refer to the video for more details.

\begin{table}
        \centering
        \caption{TEST RESULT}
        \label{table2}
        \begin{tabular}{|c|c|c|c|c|}
                \hline
                \multirow{2}{*}{ENV} & \multirow{2}{*}{METHOD} & \multicolumn{3}{c|}{METRICS}\\ \cline{3-5}
                \multirow{2}{*}{} & \multirow{2}{*}{} & Ar & Aav & Atd \\ \hline
                \multirow{3}{*}{Env. a} & sensor-level & 0.723 & 0.636 & 4.889 \\ 
                                        & map-based & 0.795 & 0.578 & 4.826 \\ 
                                        & angular map & 0.757 & 0.569 & \textbf{4.697} \\
                                        & \textbf{logarithmic map} & \textbf{0.905} & \textbf{0.5304} & 5.025 \\\hline
                \multirow{3}{*}{Env. b} & sensor-level & 0.884 & 0.571 & 15.100 \\ 
                                        & map-based & 0.893 & 0.558 & 19.073 \\
                                        & angular map & 0.885 & 0.526 & 19.806 \\
                                        & \textbf{logarithmic map} & \textbf{0.913} & \textbf{0.515} & \textbf{13.952} \\ \hline
                \multirow{3}{*}{Env. c} & sensor-level & 0.626 & 0.610 & 13.949 \\ 
                                        & map-based & 0.626 & 0.643 & 19.210 \\ 
                                        & angular map & 0.746 & 0.608 & 17.518 \\
                                        & \textbf{logarithmic map} & \textbf{0.808} & \textbf{0.523} & \textbf{11.667} \\  \hline
                \multirow{3}{*}{Env. d} & sensor-level & 0.620 & 0.685 & 6.141 \\ 
                                        & map-based & 0.815 & 0.614 & 5.179 \\ 
                                        & angular map & 0.600 & 0.656 & \textbf{4.923} \\
                                        & \textbf{logarithmic map} & \textbf{0.940} & \textbf{0.606} & 5.178 \\  \hline      
                \multirow{3}{*}{Env. e} & sensor-level & 0.533 & 0.681 & 6.308 \\ 
                                        & map-based & 0.553 & 0.674 & 6.521 \\ 
                                        & angular map & 0.607 & 0.634 & \textbf{6.284} \\
                                        & \textbf{logarithmic map} & \textbf{0.860} & \textbf{0.628} & 6.380 \\  \hline                                        
        \end{tabular} 
\end{table}

\begin{figure}
	\centering
        \subfigure[Trajectory of robot in the static scenario]{
                \label{fig9a}
                \includegraphics[ width=0.9\linewidth]{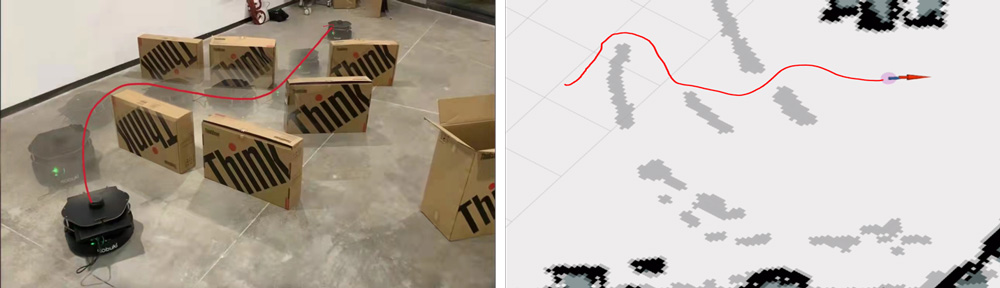}}  
        \subfigure[Trajectory of robot in in the fast scenario]{
                \label{fig9b}
                \includegraphics[ width=0.9\linewidth]{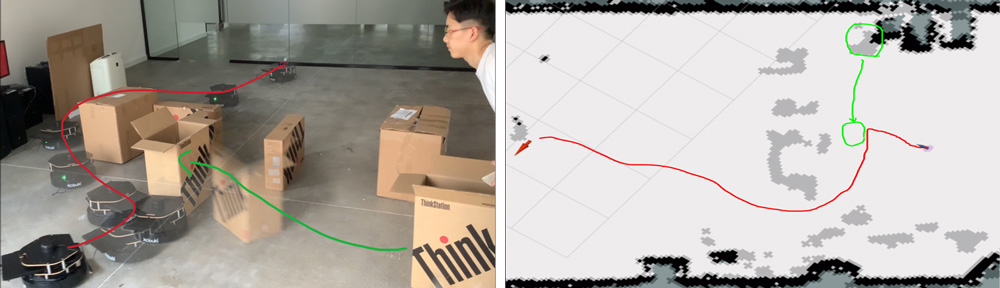}} 
        \subfigure[Trajectory of robot in in the blocked scenario]{
                \label{fig9c}
                \includegraphics[ width=0.9\linewidth]{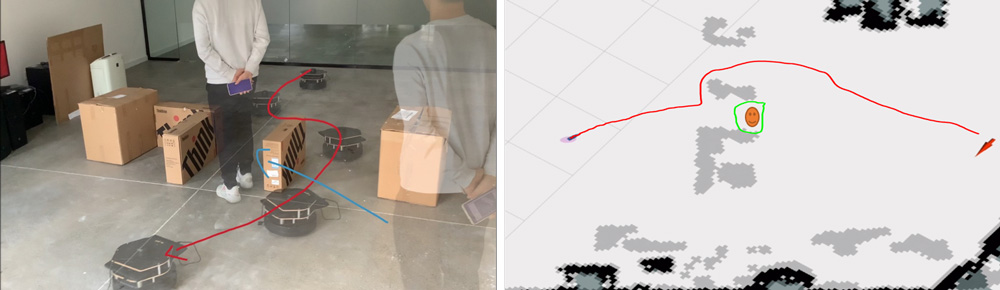}} 
        \subfigure[Trajectory of robot in in the pedestrian scenario]{
                \label{fig9d}
                \includegraphics[ width=0.9\linewidth]{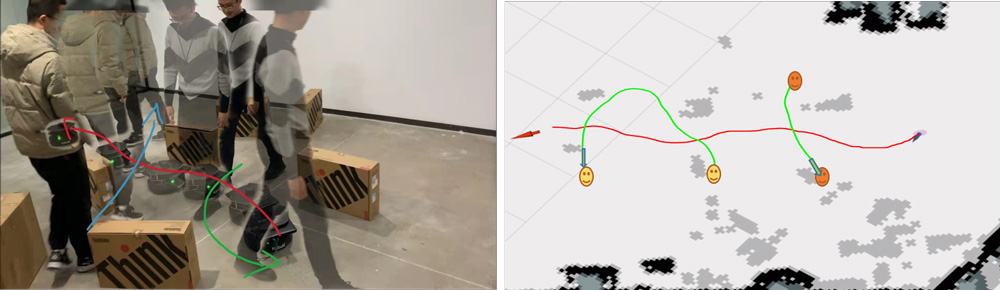}} 
	\caption{The purple circle and the blue arrow  in the figure  indicate the starting position and orientation of the robot respectively, and the red arrow represents the endpoint.}	\label{fig9}
\end{figure}

\begin{figure*}[htbp]
	\centering
        \subfigure[Trajectories of two robots in the corridor scenario.]{
                \label{fig10a}
                \includegraphics[ width=0.45\linewidth]{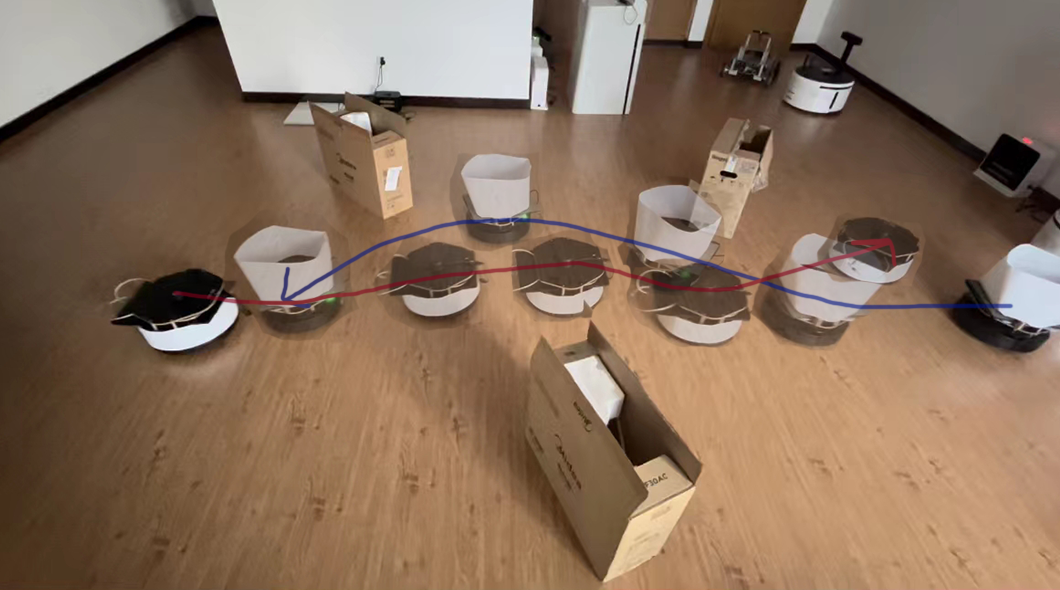}}  
        \subfigure[Trajectories of two robots in the corridor scenario with one moving people.]{
                \label{fig10b}
                \includegraphics[ width=0.45\linewidth]{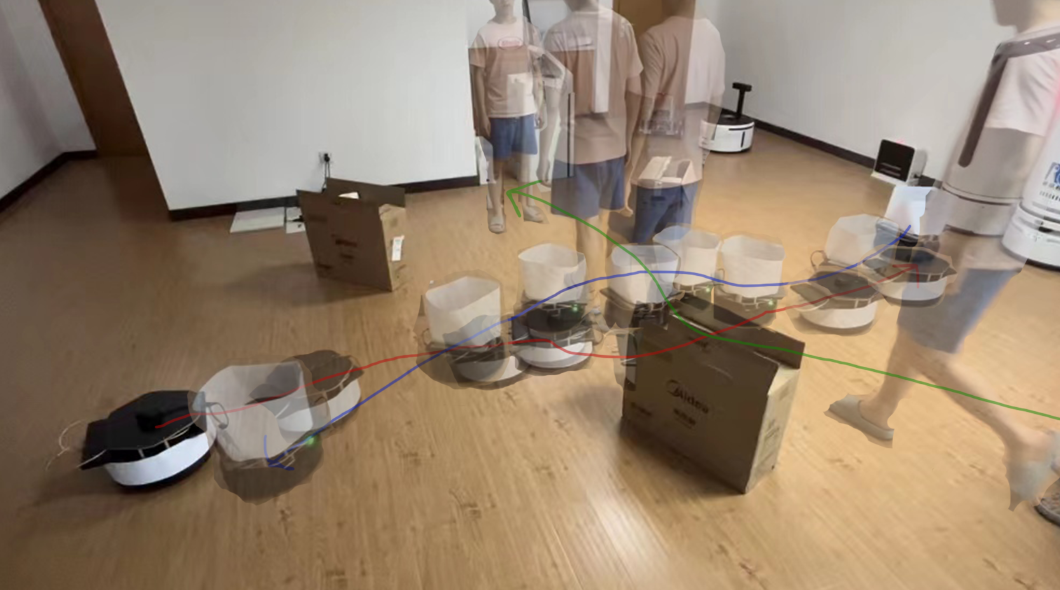}} 
	\caption{The white paper surrounded by the robot is used to help the robot perceive each other via laser scan.}	\label{fig10}
\end{figure*}

\subsection{Real-world Experiments}
\label{reality}
In the real experiment, our hardware includes turtlebot2, rplidar A2, laptops with i5-7300HQ CPU and NVIDIA GTX 1060 GPU. We design six real-world scenarios to demonstrate the performance of our model, namely:
\begin{itemize}
        \item Static scenario: a scenario with some static obstacles;
        \item Fast scenario: a scene in which fast-moving obstacles suddenly appear;
        \item Blocked scenario: a scenario where the passage is blocked suddenly.
        \item Pedestrian scenario: a scenario with two pedestrians.
        \item Corridor scenario: two robots moving in opposite directions swap their positions via a narrow corridor.
        \item Corridor scenario with moving people: a corridor scenario with one moving people.
\end{itemize}

We apply the model trained in Comb2 in a real robot as the real environment is crowded and narrow. 

Fig.~\ref{fig9} shows the test results and the corresponding trajectories of one robot in different kinds of scenarios. In Fig.~\ref{fig9a} the robot navigates in a static complex scene without collision. In Fig.~\ref{fig9b} we throw a box in front of the robot, and the robot stops in time and turns left to continue its task. In Fig.~\ref{fig9c} when the robot is about to pass through a passage, a man suddenly blocks the passage, even if the original path is blocked, the robot can quickly select alternative routes. In Fig.~\ref{fig9d} two people walk around the robot, and the robot can avoid pedestrians and complete navigation tasks.

Fig.~\ref{fig10} shows the test results and the corresponding trajectories of two robots in two kinds of scenarios. In Fig.~\ref{fig10a} two robots moving in opposite directions swap their positions in a static scenario without collision. In Fig.~\ref{fig10b} a person moves in the scenario while two robots moving in opposite directions swap their positions.

Experiments show the robot responds quickly to moving dynamic obstacles and can successfully avoid obstacles. In addition, even if there is no pedestrian element in the training process, the robot can avoid pedestrians safely.
Please refer to our online video for the specific performance of the robot.

\section{CONCLUSIONS}
\label{conclusion}

In this paper, we propose a logarithmic map-based reinforcement learning multi-robot obstacle avoidance method that can help robots focus nearby and avoid obstacles efficiently. 
We train our agents in two types of scenarios and test them in more complex scenarios. With the help of the logarithmic map, the robot can make a rapid response strategy to the obstacles around, so that it can have a satisfactory obstacle avoidance ability in a crowded and narrow environment. Besides, our method is easy to deploy on physical robots, 
and the real robot experiments also show that our method is more stable and efficient while moving in complex scenarios 
by using the logarithmic map.


\end{document}